\documentclass[letterpaper]{article}
\usepackage{aaai25} 
\usepackage{times} 
\usepackage{helvet} 
\usepackage{courier} 
\usepackage[hyphens]{url} 
\usepackage{graphicx} 
\usepackage{graphicx}  
\usepackage{natbib}  
\usepackage{caption}  
\usepackage{listings}
\title{Learning Lifted STRIPS Models from Action Traces Alone: \\ A Simple, General, and Scalable Solution}

\author{Jonas Gösgens, Niklas Jansen, Hector Geffner}

\affiliations{RWTH Aachen University, Germany}

\usepackage{booktabs}
\usepackage{enumerate}
\usepackage{xspace}
 \usepackage{todonotes}
\usepackage{amssymb}
\usepackage{amsmath}
\usepackage{amsthm}
\usepackage{mathtools}
 \presetkeys{todonotes}{inline}{}

\usepackage{algorithm,algpseudocode}

\newtheorem{definition}{Definition}
\newtheorem{theorem}[definition]{Theorem}

\newcommand{\tup}[1]{\langle #1 \rangle}

\theoremstyle{definition}

\newtheorem{assumption}{Assumption}
\theoremstyle{remark}

\theoremstyle{plain}

\usepackage{lipsum}

\newcommand{\sift}{\textsc{sift}\xspace}

\DeclareMathOperator\seconds{s}

\begin{document}

\maketitle

\begin{abstract}
Learning STRIPS action models from action traces alone is a challenging
problem as it involves learning the domain predicates as well. In this work,
a novel approach is introduced which,  like the well-known  LOCM systems,
is scalable, but  like SAT approaches, is  sound and complete. 
Furthermore, the approach is general and imposes no restrictions on the hidden domain or the number or
arity of the predicates. The new learning  method is based  on an \emph{efficient, novel test}
that checks whether the assumption that a predicate is 
affected by a set of action patterns,  namely, actions with specific argument positions,
is consistent with the traces. The predicates and action patterns that pass the test provide the basis for the
learned domain that is then easily completed with preconditions and static predicates.
The new method is studied theoretically and experimentally. For the latter, the method is
evaluated on traces and graphs obtained from standard classical  domains like the 8-puzzle,
which involve hundreds of thousands of states and transitions. The learned representations are
then verified on larger instances.
\end{abstract}

\section{Introduction}

The problem of learning lifted STRIPS action models from action traces alone is challenging
because  it requires learning the domain predicates which are not given. 
The problem has been addressed by  the LOCM systems \cite{locm1,locm}, and more recently,
through  SAT and  deep learning approaches \cite{bonet:ecai2020,ivan:kr2021,asai:latplan,asai:jair}.
Still  all these approaches have severe limitations. The LOCM systems are scalable but heuristic, and their
scope is not clear and can fail even in simple domains. The SAT approaches are  sound and complete, 
but they work on graphs,  not traces, and more importantly, they do not scale up.
Deep  learning approaches 
can deal with traces combining state images
and actions, but do not yet  produce meaningful lifted representations. 

In this work, we address the problem 
using a different approach that combines the benefits of some of  these methods while avoiding their pitfalls.
Like  LOCM, the new method is scalable, and like SAT approaches, it is  sound and complete,
while learning from either  action traces or graphs,  without
making any  assumptions on  the ``hidden'' STRIPS domains except that
they are  ``well formed'', so that action effects must change the state.

The new method, named \sift, utilizes an \emph{efficient, novel test} that checks whether it is  \emph{consistent
	with the inputs (traces or graphs) to assume that a predicate can be affected by a set of action patterns};
namely, actions with specific argument positions. The predicates and action patterns that pass the test
provide the basis for the learned domain that is then easily completed with preconditions and static predicates.
The approach is evaluated on traces and graphs obtained from standard  domains like the 8-puzzle, which involves
hundreds of thousands of states and transitions, and the learned domains are verified on larger instances.

The rest of the paper is organized as follows: We preview  the new ideas through an example, discuss related work, 
review background notions, and  introduce the  new learning formulation.  Then we present details of the implementation,  experimental results, and a summary and discussion.

\section{Preview}

The intuition for the approach is simple. Consider for example the following action trace $\tau$ for the Delivery domain:
\[pick(o_1,c),move(c,c'),drop(o_1,c'),pick(o_1,c') \ .  \]
This is an applicable  sequence of  actions in an instance of the domain
where an object $o_1$ is picked from a cell $c$, a move is done from $c$ to  cell $c'$,
the object is dropped, and picked up again.
The task is to learn the hidden domain from traces such as this, and this involves
learning the  predicates and action schemas for the three  actions
$pick$, $drop$, and $move$ including their effects and preconditions.

Given traces like $\tau$  drawn from a hidden domain $D$,
the new  method will address the learning task by considering questions like the following,
where the symbol ``\_'' stands for any values (``don't cares''): 

\begin{enumerate}
	\item Is the assumption that $D$  involves a unary atom $p(x)$  affected only  by actions of the form $pick(x,\_)$ and $drop(x,\_)$,  \emph{consistent}  with $\tau$?
	\item Is the assumption that $D$ involves  a unary atom $q(x)$ affected only by actions $pick(x,\_)$, \emph{consistent} with $\tau$?
\end{enumerate}

We will show  that these questions can be answered in a crisp and efficient manner, and moreover, that
a    domain equivalent  to the one generating the traces  can be  obtained from the answers to such questions.

\section{Related Work}

The problem of learning lifted STRIPS models has a long history in planning. Most works however have been focused on learning
lifted action models from traces that combine actions and states where the \emph{domain predicates are given}.
Observability of  these traces  can be partial, complete, or noisy
\cite{yang2007learning,mourao2012learning,zhuo2013action,arora2018review,aineto2019learning,lamanna2021online,verma2021asking,le2024learning,balyo2024learning,bachor2024learning,xi2024neuro,aineto2024action}. Fewer  works have  considered  inputs that are pure action traces.

\medskip

\noindent \textbf{LOCM:} The work that is closest to ours is the one on the LOCM system \cite{locm,locm1,locm2,locm3}, which
also accepts action traces as inputs, and outputs lifted domain descriptions. Moreover, there is a common intuition guiding
both works, namely, that the  information to be extracted from the action traces is about the  action that are ``consecutive''
in a trace. In LOCM, this happens when two ground actions in a trace share a common object as argument,
and no ground action between them does. In our setting, this basic intuition is formulated in a  different way.
The notions of action patterns and features provide the basis of  \sift, which is scalable like LOCM but with
the right theoretical properties and with no domain  restriction.
The LOCM approach, on the other hand,  is heuristic and does not have a clear scope where it is  sound and complete.{\footnote{
		Using  terminology to be introduced  later, LOCM can be thought as learning nullary and unary features (predicates),
		but not all of those which are required. \sift, on the other hand, is complete and has no arity restrictions.
		A simple example with nullary predicates only which is  beyond the reach of LOCM  is the following: actions
		$a: \neg r \rightarrow r$, $b: r, \neg p_1 \rightarrow \neg r, p_1$, $c: r, \neg p_2 \rightarrow \neg r, p_2$,
		$d: p_1, p_2 \rightarrow \neg p_1, \neg p_2$, where $a: A \rightarrow B$ stands for action $a$ with preconditions $A$ and effects $B$. 
		LOCM  will identify $p_1$ or $p_2$ but not both, because one $p_i$ is sufficient to explain  why   $d$ cannot be done  twice in a row.
}

\medskip

\noindent \textbf{SAT:} A very different approach has been aimed at learning domains  from
labeled state graphs $G$ of hidden domain instances.  For this, the simplest
instance that produces the structure of graph  $G$ is sought.
The  problem is addressed with weighted Max-SAT solvers \cite{bonet:ecai2020}
and ASP solvers \cite{ivan:kr2021}. The limitations of the approach
is that it does not learn from traces  and does not scale up to graphs
with more than a few hundred states.

\medskip

\noindent \textbf{Deep learning:} \textsc{latplan}  learns  STRIPS models without supervision from traces where 
states are represented by images \cite{asai:latplan,asai:jair}. For this, an encoder mapping states into
latent representations is learned along with a model that predicts the next latent representation
given the action. This enables  planning in latent space from new initial states encoded by images.
The approach, however, is propositional and hence does not  generalize to different state spaces,
while  approaches that aim to so, do not result yet  in meaningful  action schemas \cite{asai:fol}.

\section{Background}

A classical STRIPS problem is a pair $P=\tup{D,I}$ where $D$ is a first-order
\emph{domain} and $I$ contains information about the instance \cite{geffner:book,ghallab:book}.
The domain $D$ has a set of predicate symbols $p$ and a set of action schemas with
preconditions and effects given in terms of  atoms $p(x_1, \ldots, x_k)$, where $p$ is a predicate
symbol of arity $k$, and each $x_i$ is an argument of the schema.
The instance information is a tuple $I=\tup{O, \textit{Init},G}$ where $O$ is a
set of object names $c_i$, and $\textit{Init}$ and $G$ are sets of \emph{ground atoms} $p(c_1, \ldots, c_k)$
denoting the initial and goal situations. A STRIPS  problem $P=\tup{D,I}$ defines a state graph $G(P)$ where nodes
represent reachable states in $P$, the root node represents the initial state, and edges indicate state transitions labeled
with the actions causing them. A path in this graph represents an action sequence
that is applicable in the state represented by the first node.

Current PDDL standards support a number of STRIPS extensions \cite{pddl:book}, 
and \sift  learns domains expressed in \emph{STRIPS  with negation} where negative literals can be used in the
initial situation,  action preconditions, and  goals.
The states in such a case are not sets of ground atoms but
sets of ground literals.  Since the goals of an instance $P=\tup{D,I}$  play no role in learning, 
we will regard $I$ as just representing the initial situation through a \emph{set of signed ground atoms} (literals).

\subsection{Traces, Extended Traces, and Graphs}

An \emph{action trace} or simply a \emph{trace} from a domain instance $P=\tup{D,I}$ is an  action sequence that is applicable from a reachable state in $P$.
An  action trace from  a domain is a trace from a  domain instance. For each action trace, there is  a hidden  initial state $s_1$ and
hidden states $s_{i+1}$ generated by the actions  in the  trace. When learning from action traces alone, these states are not known, and
moreover,  there is no assumption about whether any pair of hidden states $s_i$ and $s_j$ represent the same state or not.
In certain cases, however, the information that two states in a trace or in different traces
represent the same state is available (e.g., traces drawn from the same state) and can be used. 
We will refer to sets of traces extended with such \emph{state equalities} as \emph{extended traces}.
Extended traces  generalize plain traces, where no state equalities are revealed, and can provide a good approximation of
labeled state graphs $G(P)$ considered in \cite{bonet:ecai2020,ivan:kr2021},  that represent
all possible traces in $P$ and all state equalities (more about this below). We refer to both  plain traces
and extended traces as traces, and make their difference explicit when relevant.

\section{Formulation}

\sift learns  lifted STRIPS model from action traces.
For this, it constructs the possible domain features
from the traces, tests their consistency, and
defines the action schemas in terms of them.
The consistent (admissible) features denote
the predicates in the domain and how they are
affected by the actions.
Figure \ref{fig:sift} visualizes this workflow.

	\begin{figure}[t]\centering
		\begin{tikzpicture}
			\node[draw] (input) {
				\begin{minipage}{0.105\linewidth}
					action\newline traces
				\end{minipage}
			};
			\node[draw, right=0.8cm of input] (candidates) {
				\begin{minipage}{0.135\linewidth}
					possible\newline features
				\end{minipage}
			};
			\node[draw, right=0.8cm of candidates] (results) {
				\begin{minipage}{0.18\linewidth}
					admissible\newline features
				\end{minipage}
			};
			\node[draw, right=0.8cm of results] (domain) {
				\begin{minipage}{0.125\linewidth}
					learned\newline domain
				\end{minipage}
			};
			\draw[->] (input) -- (candidates);
			\draw[->] (candidates) -- (results);
			\draw[->] (results) -- (domain);
		\end{tikzpicture}
		\caption{\sift: Model learning from action traces}
		\label{fig:sift}
	\end{figure}

An  assumption in \sift is that the observed action traces
come from a hidden STRIPS domain with negation which is \emph{well-formed}:

\begin{assumption}
	The hidden  STRIPS domain $D$ is \emph{well-formed} in the sense
	that  if an   action  makes a literal true,  the  complement   of the literal
	is an action precondition.
\end{assumption}

This assumption  rules out situations where an action adds a literal
that is   already  true, or deletes a literal that is  already false.
The assumption can be enforced by including the complement of effects
as preconditions. The result is that action effects always change the state.

\subsection{Dual Representation of Action Effects}

While a  planning domain $D$ is normally described in terms of 
a set of \emph{actions schemas} with their preconditions and effects,
there is an alternative  way of describing \emph{action effects}
without their signs, that will be  convenient in our setting.
For example, a domain $D$ with two actions schemas
\begin{align*}
	&   \hbox{Action } a(x_1,x_2):  \, \hbox{ Effects } p(x_1,x_2), \\
	&   \hbox{Action } b(x_1,x_2,x_3): \,  \hbox{Effects }  q(x_1,x_2),  q(x_3,x_2)
\end{align*}

\noindent can  also be  expressed in term of the atoms  affected by the actions as:
\begin{align*}
	&  \hbox{Atom }  p(x_1,x_2): \, \hbox{Affected by }  a(x_1,x_2) \\
	&  \hbox{Atom }  q(x_1,x_2): \, \hbox{Affected by }  b(x_1,x_2,\_),  b(\_,x_2,x_1) , 
\end{align*}

\noindent where the missing arguments ``$\_$'' are don't cares that can take any
value. We will refer to lifted actions of the form $b(y_1,y_2,\_)$
and $b(\_,y_1,y_2)$, as \emph{action patterns}.

In the normal representation of  action effects, \emph{each action schema occurs once
	and may involve many atoms with the same predicate}; in the alternative, 
dual representation,  \emph{each lifted atom occurs once and may involve multiple
	action patterns}. The two representations  are equivalent.

\subsection{Action Patterns and Features}

We will say that an action $b(x_1,x_2,x_3)$
has the atom $q(x_3,x_2)$ as an effect, by saying that
the predicate $q$ is affected by the \emph{action pattern} $b[3,2]$.
This means that  the arguments of $q$ bind to the third and second arguments
of $b$ respectively.  Formally, an action pattern is:

\begin{definition}[Action patterns]
	An \emph{action pattern}  $a[t]$ of arity $k$ 
	is an action name $a$ of arity $k' \ge  k$ followed by a tuple $t$
	of $k$ different indexes $t= \tup{t_1, \ldots, t_k}$, $1 \leq t_i \leq k'$, $i=1, \ldots, k$.
\end{definition}

A  lifted  atom in a domain  is  not affected by  actions $a$
but by \emph{action patterns} $a[t]$ that bind  the arguments of the
atom to the action arguments. For learning a domain,
leaving  preconditions and the sign of effects aside,
it will be  sufficient  to learn the predicates $p$ involved in the domain and
the \emph{action patterns} $a[t]$ that affect them. We will refer to
the predicates that are possible given a set of action patterns, as \emph{features}:

\begin{definition}[Features]
	A \emph{feature} $f$ of arity $k$ is a pair $f=\tup{k,B}$, where $B$ is a non-empty
	set of action patterns of  arity $k$. $B$ is  called  the \emph{feature support}, also referred to
	as $B_f$.
\end{definition}

A feature $f=\tup{k,B}$ represents an  \emph{assumption} about the hidden domain;
namely, that it contains   atoms   $f(x_1, \ldots, x_k)$ of arity $k$
which are affected by \emph{all and only} the action patterns $a[t]$  in $B$
which must have the same arity $k$. The actions $a$ in these patterns, however,
can have any arity $k'$ greater than or equal to  $k$, as the indexes in the pattern $t$  select the $k$
relevant action arguments of $a$ in order. For example, if $k'=3$,  
the action  pattern $a[3,2]$ in $B$ says  that $f(x_3,x_2)$ is an effect of the action $a(x_1,x_2,x_3)$.
A finite set of inputs in the form of action traces determines a finite  set of action names $a$ with their arities,
and these  determine in turn a finite set of action patterns $a[t]$ and a finite set of features. 
The learning task will  reduce  to a large extent to finding the features that are consistent with the  given
traces.

\subsection{Action Groundings $A_f(o)$}

The \emph{action grounding} $A_f(o)$ of a feature $f$ in a given set $T$ of traces, 
will  refer to the set of ground actions $a(o')$ in $T$ that are assumed to  affect 
the truth of the hypothetical ground atom $f(o)$. 
For making this precise, for an action pattern $a[t]$, $t=\tup{t_1, \ldots, t_k}$,
and a  ground action $a(o)$, $o=\tup{o_1, \ldots, o_{k'}}$,  $k' \ge k$,
let $t_i[o]$ refer to $o_j$ if $t_i=j$, and let $t[o]$ refer to the tuple
of objects $t_1[o], \ldots, t_k[o]$. Then the action grounding $A_f(o)$
can be defined as follows:

\begin{definition}[Action groundings]
	The \emph{action grounding} $A_f(o)$ of a feature $f$ in a set of traces $T$
	refers  to the set of ground actions $a(o')$ in $T$ such that $a[t]$ is a pattern
	in $B_f$ and $o=t[o']$.
\end{definition}

For example, if $o=\tup{o_1,o_2}$ and $a[4,1]$ is a  pattern in $B_f$,
$A_f(o)$ will include all the  ground actions in $T$ of the form
$a(o_2, \_, \_, o_1)$, as $o=t[o']$ is true for $t=[4,1]$
if the two  elements of $o$ are the fourth and first elements of $o'$.

The action grounding $A_f(o)$ contains all and only the ground actions appearing in $T$ that 
affect the truth of the hypothetical atom $f(o)$. 
Interestingly,  by just ``looking'' at the traces in $T$, we will be able to tell in time
that is linear in the  length of the traces, whether the assumption expressed by a feature
is consistent with the traces.

\subsection{Pattern Constraints}

Each  pattern  $a[t]$ in $B_f$ represents an effect on the hypothetical atom $f(x)$,
which  must hence  have a \emph{sign}:  positive ($1$) when the  atom becomes true,
and negative ($0$) when the atom becomes false. 

\begin{definition}[Signs]
	Each pattern $a[t]$ in $B_f$ has a   unique sign, $sign(a[t])$,  that can be 0 or 1.
\end{definition}

A feature $f=\tup{k,B}$ is consistent with the input traces if
it's possible to assign a sign to each action pattern $a[t]$ in $B$
in a way that is compatible with the traces.
For this, we extract two types of  \emph{pattern constraints}  from sets of traces $T$:
those that follow from patterns appearing sequentially in some grounding of $T$,
and those that follow from  patterns appearing in parallel in some grounding of $T$.

\begin{definition}[Consecutive patterns]
	Two action patterns $a[t]$ and $b[t']$ in $B_f$ are \emph{consecutive} in $T$, 
	if  some trace $\tau$ in $T$ contains two actions $a(o_1)$ and $b(o_2)$
	that appear in some action  grounding $A_f(o)$, such that 
	no other action from  $A_f(o)$ in the trace appears between them.
\end{definition}

If  $a[t]$ and $b[t']$  affect  $f$ and are \emph{consecutive} in $T$,
the two patterns must have different signs. Indeed if $a(o_1)$ 
adds the atom $f(o)$, $b(o_2)$ must delete $f(o)$, and vice versa,
if $b(o_2)$ adds it, $a(o_1)$ must delete it. 

A different type of constraint  on action patterns arises from (extended) traces
that start at or reach the same state:

\begin{definition}[Fork patterns]
	Two patterns $a[t]$ and $b[t']$ in $B_f$ express a  \emph{fork} in a set of extended
	traces $T$ if actions  $a(o_1)$ and $b(o_2)$ in  $A_f(o)$ appear
	in traces $\tau_1$ and $\tau_2$ respectively, that diverge from (resp. converge to)  the same
	state, and no other action in $A_f(o)$  appears between the common state and each of the actions
	(resp. between each of the actions and the common state).
\end{definition}

If the action  patterns $a[t]$ and $b[t']$ in $B_f$ express a   \emph{fork}  in $T$ arising from a state $s$
or converging to a state $s'$, they will have the same $f(o)$ precondition in $s$ or the
same $f(o)$ effect in $s'$, and in either case, {they must have the same sign.}

The set of constraints resulting from a set of plain or extended traces is the following: 

\begin{definition}
	The set of $C_f(T)$ of pattern  constraints is given by the  \emph{inequality constraints}   $sign(a[t]) \not= sign(b[t'])$ for
	\emph{consecutive patterns} $a[t]$ and $b[t']$ from $B_f$ in $T$, and by the \emph{equality constraints} $sign(a[t]) = sign(b[t'])$ for 
	\emph{fork patterns}  $a[t]$ and $b[t']$ from $B_f$ in $T$.
\end{definition}

Two ``border conditions''  need to be handled as well.  We  will also  regard  two   patterns $a[t]$ and $b[t']$ in $B_f$  as 
a  \emph{fork} if  there is an action in some grounding $A_f(o)$   which  is an instance of the two patterns (for this, $a=b$),
and as \emph{consecutive}  patterns if instances of $a[t]$ and $b[t']$  are the  first and last action in a grounding  $A_f(o)$  of
a  (sub) trace  that connects  two states which  are  connected by another (sub) trace(s) with an empty grounding $A_f(o)$.
In the first case, the patterns $a[t]$ and $b[t']$ in $B_f$ must have the same sign because they account for the effect of the
same ground action over $f(o)$, while in the second case, the patterns must have different signs because the same $f(o)$ literal
must be a precondition and negative effect of the first action, and a positive effect of the second.

\section{Feature Consistency}

The constraints $C_f(T)$ for the feature $f$ are extracted from the given traces $T$, and the solution to these constraints
is a \emph{sign assignment} to the patterns $a[t]$ in $B_f$; namely, a 0/1 valuation  over the expressions $sign(a[t])$, 
such that all  the constraints in $C_f(T)$ are satisfied. If there is one such valuation, the set of constraints $C_f(T)$
and the feature $f$ are said to be \emph{consistent}, and the sign of the patterns $a[t]$ in $B_f$ is given by such a valuation.

\begin{definition}[Feature consistency]
	The feature $f$ is \emph{consistent} with a  set of (extended) traces $T$ if the set of pattern contraints $C_f(T)$ is \emph{consistent.}
\end{definition}

The good news is that both the  extraction of  the pattern constraints from traces, and the  consistency test
are easy computational problems. The latter can indeed be reduced to 2-CNF satisfiability:

\begin{theorem}
	The problem of determining if a feature $f$ is consistent with a set of (extended) traces $T$ is in $P$
	and reduces to the problem of checking \emph{2-CNF satisfiability.}
	\label{theorem:2-CNF-easy}
\end{theorem}

The reduction is direct: if the propositional symbol $p_{a[t]}$ stands for $sign(a[t])=1$, then
the equalities  $sign(a[t]) = sign(b[t'])$ map into  implications  $p_{a[t]} \rightarrow p_{b[t']}$
and $\neg p_{a[t]} \rightarrow \neg p_{b[t']}$, and  inequalities  $sign(a[t]) \not = sign(b[t'])$
into implications $p_{a[t]} \rightarrow \neg p_{b[t']}$ and $\neg p_{a[t]} \rightarrow  p_{b[t']}$,
all of which define clauses with two literals. A 2-CNF formula is  \emph{unsatisfiable} iff
implication chains  $p \rightarrow l_1 \rightarrow l_2  \cdots \rightarrow  \neg p$ and
$\neg p \rightarrow l'_1 \rightarrow l'_2  \cdots \rightarrow p$ can be constructed for one of
the symbols $p$. For our constraints,  the first chain implies the second, and vice versa,
so that the satisfiability algorithm required  is even simpler. 
For checking the  consistency of a  feature $f$ given the traces $T$, an arbitrary pattern $a[t]$ in $B_f$
is chosen and  given the arbitrary  value $1$. Then,  all patterns   $b[t']$ in $B_f$ that are directly related to $a[t]$
through  a  constraint in $C_f(T)$ and which  have no value, get the same value as $a[t]$, if the relation is equality,
and the inverse value if the relation is inequality. If there  are then  patterns in $B_f$  that did not get a value,
one such pattern $a[t]$ is chosen  and given  value $1$,  and the whole process is repeated  over such patterns.
The iterations continue til an inconsistency is detected or \emph{all patterns get a sign}.
The algorithm runs in time that is linear in the number of patterns in $B_f$.

\bigskip

\noindent \textbf{Example}. We considered  the trace $\tau$ given by the
action sequence  $pick(o_1,c)$, $move(c,c')$, $drop(o_1,c')$, $pick(o_1,c')$. 
The question was whether  a unary atom $p(x)$ affected by  actions of the form
$pick(x,\_)$ and $drop(x,\_)$ is    {consistent}  with the trace. The question becomes
now whether the  feature $f = \tup{1,B}$ with $B=\{pick[1],drop[1]\}$ is consistent with
the set of traces $T=\{\tau\}$. For this, the only (non-empty) action grounding $A_f(\tup{o_1})$
in $\tau$ for $f$ is given  by the  set of  actions $pick(o_1,c)$, $drop(o_1,c')$, and $pick(o_1,c')$. There are two
pattern \emph{inequality constraints}  then in $C_f(T)$  that follow from pattern $drop[1]$ following pattern $pick[1]$ in $\tau$, and $pick[1]$ following  $drop[1]$.
These consecutive patterns in $B$ result in the  single constraint  $sign(pick[1]) \! \not= \! sign(drop[1])$,
which is indeed satisfiable, so feature $f$ is \emph{consistent} with the trace $\tau$.
On the other hand,  the other feature considered, $f' = \tup{1,B'}$ with $B'=\{pick[1]\}$,
is \emph{not consistent} with $T$ as the the action grounding $A_{f'}(\tup{o_1})$  contains just  the two  $pick(o_1,\_)$ actions
in $\tau$ (the other actions not being in $B'$), with one $pick[1]$-pattern following  another $pick[1]$-pattern,
resulting in the unsatisfiable pattern constraint $sign(pick[1]) \not= sign(pick[1])$.

\section{From Features to  Domains}

We will refer to features that are consistent with the given traces $T$  as   \emph{admissible}, 
and to the collection of admissible features,  as   $F(T)$. We show first how to use these  features
to define the  learned domains $D_T$ and the  learned instances $P_T=\tup{D_T,I_T}$ from $T$. 
For this, notice that  a trace in $T$ with a non-empty  action grounding  $A_f(o)$,
defines a \emph{unique truth value}, up to negation, for the atom
$f(o)$ in \emph{every state of the trace}. Likewise, in a set of \emph{connected (extended) traces} where each pair of traces
shares a common state, a non-empty grounding $A_f(o)$ in one of the traces, determines a \emph{unique truth value} 
for the atom $f(o)$, up to negation, in every state of all of the connected  traces.\footnote{
	A trace with a  non-empty action grounding $A_f(o)$  defines the  truth value of $f(o)$ right after and right before any action
	$a(o') \in A_f(o)$; the first is  the  sign of the pattern $a[t]$ in $B_f$ for which $o=t[o']$, the latter is the inverse sign.
	These values persist along the trace, forward after $a(o')$ and backward before $a(o')$ til another action
	$b(o'') \in A_f(o)$ appears in the trace, if any. Such an action inverts the value of the atom, and the
	propagation continues in this way til reaching the beginning and end of the trace.} These truth values are used
to infer the  action preconditions in the learned domain $D_T$:

\begin{definition}[Learned domain]
	The \emph{domain} $D_T$ learned from a set of action traces $T$ is defined as follows:
	\begin{itemize}
		\item Lifted actions $a(x)$ with arities as  appearing in  $T$.
		\item Predicates $f(x)$ of arity $k$ if $f=\tup{k,B}$ is in $F(T)$.
		\item Effects $f(t[x])$ of $a(x)$ with $sign(a[t])$ if $a[t] \in B$.
		\item Preconditions  $f(t[x])$ (resp. $\neg f(t[x])$) of $a(x)$ for $f \in F(T)$,
		if there are  traces  where an action $a(o')$ is applied for $o=t[o']$,
		and in all such traces   $f(o)$ is true (resp. false)  right before $a(o')$.
	\end{itemize}
	\label{def:domain}
\end{definition}

The expression $t[x]$ selects elements of $x$ according to the indices in $t$,
and while $t$ in  \emph{effects} comes from the action pattern $a[t] \in B$;
$t$ in \emph{preconditions} ranges over the possible \emph{precondition patterns of $a$}. That is, 
if the arities of $f$ and $a$ are $k$ and $k' \leq k$,  then $t$ ranges over all tuples $\tup{t_1,\ldots, t_k}$
where the indexes $t_i$ are different and  $1 \leq t_i \leq k$.  The reason that action  preconditions can be learned by just  taking the ``intersection''
of the $f(t[x])$-literals  that are true when the action $a$ is applied,  is that such literals
include the true, hidden, domain literals, as we will see in the next section.

The  instance $P_T=\tup{D_T,I_T}$ \emph{learned} from a set of \emph{connected traces} $T$ is
defined in terms of  the set $A(D,T)$ of ground atoms $f(o)$ \emph{whose truth values
	over all  states along the traces in $T$} are  determined by the domain $D$ and the traces  $T$.
These are:

\begin{definition}[Relevant ground atoms]
	$A(D,T)$ stands for the set of   ground atoms  $p(o)$ 
	such that $p$ is a predicate in $D$, and   some action $a(o')$ in a trace in $T$
	has an effect or  precondition $p(o)$ in $D$ with any sign.
\end{definition}

\noindent Indeed, the truth values of the atoms $p(o)$ in $A(D,T)$ in each of the
states over the traces in $T$, follow from $D$ and $T$ by simple constraint propagation:

\begin{theorem}[Truth values]
	The truth values of each ground atom $p(o)$ in $A(D,T)$ in each of the states  $s$ 
	underlying a set of connected  traces in $T$ are fully determined by $D$ and $T$.
	\label{theorem:Values-Relevant-Atoms-determined}
\end{theorem}

The truth values are determined because the signs of the action preconditions and effects
in $D$ are known,  and every atom $p(o)$ in $A(D,T)$ is the precondition or effect of an action in a trace from $T$.
The  instance $P_T=\tup{D_T,I_T}$ learned from $T$ assumes that  at least one trace from $T$
is  drawn from the initial state of  $P=\tup{D,I}$. We call the initial
state of this trace, the initial state of $T$:

\begin{definition}[Learned instance]
	For a set of \emph{connected traces} $T$ drawn from a hidden instance $P=\tup{D,I}$,
	the \emph{learned  instance} is $P_T=\tup{D_T,I_T}$ where  $D_T$ is the domain learned from $T$, and 
	$I_T$  is $A(D_T,T)$ with the  truth values of the atoms $f(o)$ in $I_T$ as derived at  the initial state of $T$. 
\end{definition}

We can now  express  a soundness and completeness result for the  instances $P_T$
learned from a   hidden instance  $P=\tup{D,I}$ with \emph{no static predicates}, as
static predicates can be treated separately (see below). For stating the conditions,
we ask the traces to be \emph{complete} in the following sense:

\begin{definition}
	A set of traces $T$  is \emph{complete} for an instance $P=\tup{D,I}$
	if the traces in $T$ are all drawn from the initial state of $P$ and hence are  connected, 
	they affect each predicate $p$ in $D$,  and  $I$ contains the same atoms as $A(D,T)$, ignoring the  signs.
\end{definition}

A set of traces $T$  is  complete for  $P=\tup{D,I}$ basically if one can infer $I$ from
the domain and the traces. The condition that the traces in $T$ affect each predicate $p$ in $D$
asks for  $p$ not to behave like a static predicate in $T$; i.e., some state is reached
by the traces where some $p$-literal changes sign.  This ensures that some feature $f \in F(T)$ will capture $p$,
as the features ``do not see'' static predicates or those that behave as such in  the traces. 
A key result is:

\begin{theorem}[Soundness and completeness]
	Let $T$ be a complete set of traces $\tau$  from   $P=\tup{D,I}$. 
	Then 1)~Each $\tau\in T$ is  applicable in the initial state of the learned  instance $P_T=\tup{D_T,I_T}$.
	2)~If $\tau$ reaches a state in $P$ where a ground action $a(o)$ that appears in $T$
	is not applicable, then  $\tau$ reaches a state in $P_T$ where $a(o)$ is not applicable. 
	\label{thm:main}
\end{theorem}

The intuition behind this result is as follows. First, the definition of the action preconditions in $D_T$,
ensures that the traces in  $T$ are executable in $P_T$, as they must all be true then. Second,  as we will see, the atoms $p(x)$
in the hidden domain $D$ define features $p$ that are admissible over any set of  traces $T$  drawn from $D$.
Thus,  if an action $a(o)$ is not applicable in $P$ after $\tau$ but  $a(o)$ is applied elsewhere in $T$, then
the truth values of the preconditions of $a(o)$ in $D_T$ will be  known in all nodes of $T$,
and   if   $T$ is complete,  such preconditions will include the true hidden preconditions of $a(o)$ in $D$.
So  hence if $a(o)$ is not applicable after $\tau$ in $P$,  it will not be  applicable after $\tau$ in $P_T$ either.

\subsection{Static Predicates}

Static predicates refer to predicates that are not changed by any action, and they just control the
grounding of the actions. Many domains are described using static predicates that detail, for example, 
the topology of a grid. Yet static predicates can be defined in a very simple and general  manner,
and while it's possible to learn them from  traces, it is not strictly necessary. For this, it suffices
to introduce a static predicate  $p_a$ for  each lifted action $a$ appearing in the traces $T$,
with the same arity as $a$. Then, for extending Theorem~\ref{thm:main} to domain $D$ with static predicates,
atoms $p_a(o)$ are set to true in $I_T$  iff the ground action $a(o)$ appears in a trace in $T$.
The theorem extends then  in a direct manner. Of course, a  more meaningful and compact  characterization of static predicates
can be obtained in terms of predicates of lower arity that may be shared across actions, but this has nothing to do
with generalization. In a new instance, the static atoms that are true initially must be given explicitly:
in one case, in terms of predicates like $p_a$, in the other case,  in terms of  predicates
of lower arity. Both forms are equally correct and the latter  is just more convenient and  concise.
Since obtaining a  compact representation of the static $p_a$ relations is not necessary, 
we will not address the problem in this work.

\section{Generalization}

We address next the relation between the hidden domain $D$ and the learned  domain $D_T$.
For this, we  make first  explicit  the notion of domain feature,
which was implicit in our  discussion of the dual representation of action effects:

\begin{definition}
	For a domain $D$ with predicates $p_1, \ldots, p_n$, the \emph{domain features}  $f_1, \ldots, f_n$ are 
	$f_i = \tup{k_i,B_i}$, where $k_i$ is the arity of $p_i$, and $B_i$ contains the action pattern $a[t^i]$
	iff the atom $p_i(t^i[x])$ is an effect of  action $a(x)$ in $D$.
\end{definition}

For example, if $a(x_1,x_2,x_3)$ has effects $p_1(x_1,x_2)$ and $p_1(x_3,x_1)$, and no other action affects $p_1$, 
the feature $f_1$ corresponding to  $p_1$ is $f_1 = \tup{2,B_1}$ where $B_1 = \{a[1,2],a[3,1]\}$. 
The first result is that features drawn from a domain $D$ are admissible given any set of  traces from $D$:

\begin{theorem}
	Let $T$ be any  set of extended traces from $D$, and let $f$ be a feature from $D$. Then, $f$ is consistent with $T$ and hence
	admissible.
	\label{thm:admissibility}
\end{theorem}

This means that in the learned domain $D_T$, there will be atoms $f(x)$  that represent the atoms in the hidden domain or their complements.
Still the learned domain $D_T$ may contain other $f(x)$ atoms as well.
There  may be indeed  many domains  $D'$ that are  equivalent to $D$, meaning that they can generate
the same set of traces and  labeled state graphs. The pairs  of equivalent domains that are interesting
are those that result in \emph{different domain features}. It turns out that the domains that are equivalent to $D$,
can be all  combined (after suitable renaming) into a single domain $D_{max}$ that is equivalent to $D$. For
this, all preconditions and effects must be joined together with their corresponding signs. The domain $D_{max}$ is equivalent to $D$
and provides indeed a \emph{maximal description} of $D$. We will refer to the  features $f$ that follow from  $D_{max}$
as the  \emph{valid features}:

\begin{definition}
	Feature $f$ is valid in $D$ if $f$ is from $D_{max}$.
\end{definition}

Since the notion of admissibility does not distinguish  $D$ from domains that are equivalent to $D$, 
Theorem~\ref{thm:admissibility}, can be generalized as follows:

\begin{theorem}
	If a feature $f$ is valid in $D$, then $f$ is consistent with any traces from $D$, and hence admissible.
	\label{theorem:All-valid-features-admissible}
\end{theorem}

A  feature $f$ that is not valid can be shown to be inconsistent in some set of extended traces,
and since there is  a finite set of features given a set of traces, this means that:

\begin{theorem}
	There is a finite set of extended traces $T$ such that $f$ is consistent with $T$ iff $f$ is valid.
	The learned domain $D_T$ is then equivalent to the hidden domain $D$.
	\label{theorem:existence-sufficient-finite-input}
\end{theorem}

Proving that a finite set of traces has this property or that a feature $f$ is valid for an arbitrary domain,
however, is  not simple and may even be undecidable in general. The experiments below  test generalization and validity
empirically.

\section{Implementation}

We explain next some relevant details about the implementation of the
domain learning algorithm  called  \sift.
The algorithm  accepts a set of traces or extended traces $T$ in the form
of graphs  with nodes  that are  hidden states and  edge labels that are actions.
For plain traces, these graphs are labeled chains. \sift  then performs
three steps: 1)~generation of the  features $f$, 2)~pruning the inconsistent
features, and 3)~construction of  the learned domain $D_T$ and of the set of
ground atoms $f(o) \in A(D,T)$ with the  truth values over  each input node.
We explained these three steps above. Here we provide details about the implementation
of  1 and 2.

\medskip

\noindent \textbf{Features.} The key idea to   make  the learning approach computationally feasible
and to avoid  the enumeration of features is the  extraction of  \emph{type information} about the  action arguments
from the traces, as done  in LOCM \cite{locm}, and its use  for making the features typed.
The types are constructed as follows.
Initially, there is a type $\omega_{a,i}$ for each action $a$ of arity $k_a > 0$ in the traces,  and each argument
index $i$, $1 \leq i \leq k_a$. Then two types $\omega_{a,i}$ and $\omega_{b,j}$ are merged into one if there is an object $o$
in the traces that appears both as the $i$-th argument of an $a$-action and as the $j$-th argument of a $b$-action.
This merging of types is iterated until a fixed point is reached, where the objects mentioned in the traces
are partitioned into a set of disjoint types. The following step is to use such  types to enumerate
the possible \emph{feature types}, and for each feature type, the possible features.
This \emph{massively} reduces the number of features $f=\tup{k,B}$ that
are  generated and checked for consistency.
We explain this through an example. In  Gripper, the actions $pick(b,g,r)$ and $drop(b,g,r)$
take three arguments of types ball, gripper, and room, while the other action,  $move(r_1,r_2)$, 
takes two arguments of  type room. Simple calculations that follow from the arities of these  actions,
show that $14$ action patterns $a[t]$ of arity two can be formed from these actions,
and thus $2^{14}-1=16,383$ features.
If types are taken into account  and read from the traces, $7$ possible binary feature types are found
(namely, ball and gripper, ball and room, etc), each of which accommodates
$2$  action patterns at most  (e.g., $pick[1,2]$ and $drop[1,2]$ for ball and gripper).
Hence, the number of (typed)  binary  features $f$ becomes $7 \times (2^2-1)=21$,
which is much smaller than $16,383$.
A further reduction is  obtained by  ordering  the  types  and using the types in the  feature arguments
in an ordering that is compatible with such a  fixed, global ordering,   avoiding the generation of
symmetrical features. This reduction leaves the number of binary features to be tested  in Gripper down to $4 \times (2^2-1)=12$.
The number of ternary (typed and  ordered)  action patterns in Gripper is 2,  and hence, there are $(2^2-1)=3$ ternary
features to check, while the number of nullary action patterns is $3$, and hence the  number of nullary features is $2^3-1 = 7$.

\medskip

\noindent \textbf{Pattern constraints $C_f(T)$.} This optimization  is critical for processing very large state graphs, not plain traces.
We will show for example that \sift  can  learn the $n$-puzzle domain  by processing the full state graph for $n=8$, which involves almost 200,000 states
and 500,000 state transitions.
For this, the  pattern constraints  $C_f(T)$ are  obtained by traversing \emph{reduced graphs} where edges $(n,n')$ labeled with actions
$a$ with no pattern in $B_f$ are eliminated by merging the nodes $n$ and $n'$. This simplification, that applies  to plain traces as well,
is carried out at the level of \emph{feature types}, because there are actions that due to  their  argument types, cannot be part of any feature
of a given type.
In our  current implementation, the process of collecting the pattern constraints in $C_f(T)$ for a given action grounding $A_f(o)$ and checking
their consistency is  done by a simple 0-1 coloring algorithm that runs in time that is linear in the size of such reduced graphs.

\begin{table*}[t]\setlength{\tabcolsep}{3.2pt}\centering
	\begin{tabular}{llll|rrrr|rrrr|rrrr}
		&                                          &&& \multicolumn{4}{c|}{Full Graphs} & \multicolumn{4}{c|}{Partial Graphs} & \multicolumn{4}{c}{Traces}\\
		Domain   & $\# O$ & $\# P$ & $\# F$ & $\# F_a$ & $\# E$& Time & Verif  & $\# F_a$ & $\# E$& Time & Verif  & $\# F_a$ & $\# E$& Time & Verif             \\
		\midrule
		blocks3      &  6 &  3 & $1220$& $5.0$&$21300 $&$   51\seconds$&$100\%$& $5.0$&$    81$&$   28\seconds$&$100\%$&  $5.0$&$    65 $&$   29\seconds$&$100\%$\\ 
		blocks4      &  7 &  5 &   $93$& $9.0$&$186578$&$  587\seconds$&$100\%$& $9.0$&$    86$&$   17\seconds$&$100\%$&  $9.0$&$    85 $&$   17\seconds$&$100\%$\\ 
		delivery     & 13 &  3 &   $62$& $5.0$&$57888 $&$  183\seconds$&$100\%$& $5.0$&$   601$&$  105\seconds$&$100\%$&  $5.0$&$   350 $&$  100\seconds$&$100\%$\\ 
		driverlog    & 11 &  4 &  $560$&$13.0$&$63720 $&$  700\seconds$&$100\%$&$13.0$&$  1201$&$  777\seconds$&$100\%$& $13.0$&$   350 $&$  683\seconds$&$100\%$\\ 
		ferry        & 10 &  4 &   $31$& $4.0$&$156250$&$  347\seconds$&$100\%$& $4.0$&$   251$&$   31\seconds$&$100\%$&  $4.0$&$   170 $&$   24\seconds$&$100\%$\\ 
		grid         & 14 &  6 &  $290$& $7.0$&$99863 $&$  546\seconds$&$100\%$& $7.0$&$ 10001$&$   37\seconds$&$100\%$& $29.7$&$ 10000 $&$   36\seconds$&$  0\%$\\ 
		grid\_lock   & 14 &  6 & $1042$& $7.0$&$152040$&$ 1678\seconds$&$100\%$& $7.0$&$   500$&$  123\seconds$&$100\%$&  $7.0$&$   800 $&$  117\seconds$&$100\%$\\ 
		gripper      & 12 &  4 &   $43$& $6.0$&$95680 $&$  212\seconds$&$100\%$& $6.0$&$   230$&$   64\seconds$&$100\%$&  $6.0$&$   250 $&$  306\seconds$&$100\%$\\ 
		hanoi        & 12 &  2 &  $134$& $4.0$&$59046 $&$ 2164\seconds$&$100\%$& $4.0$&$    50$&$   25\seconds$&$100\%$&  $4.0$&$    25 $&$   24\seconds$&$100\%$\\ 
		logistics    & 18 &  2 &  $212$& $7.0$&$648648$&$12866\seconds$&$100\%$& $7.0$&$  5003$&$ 1403\seconds$&$100\%$&  $7.0$&$   350 $&$ 1304\seconds$&$100\%$\\ 
		miconic      & 10 &  3 &   $99$& $8.0$&$127008$&$  376\seconds$&$100\%$& $8.0$&$    46$&$   33\seconds$&$100\%$&  $8.0$&$    60 $&$   36\seconds$&$100\%$\\ 
		npuzzle      & 14 &  2 &  $912$&$26.0$&$483840$&$11598\seconds$&$100\%$&$26.0$&$   130$&$  312\seconds$&$100\%$& $26.0$&$   200 $&$  311\seconds$&$100\%$\\ 
		sokoban      & 16 &  2 &  $352$& $3.0$&$26834 $&$  231\seconds$&$100\%$& $3.0$&$ 10002$&$  195\seconds$&$100\%$&$126.2$&$ 10000 $&$  220\seconds$&$  4\%$\\ 
		sokoban\_pull& 16 &  2 &$20740$& $3.0$&$66328 $&$  401\seconds$&$100\%$& $3.0$&$   300$&$   77\seconds$&$100\%$&  $3.0$&$   300 $&$  123\seconds$&$100\%$\\ 
		
	\end{tabular}
	\caption{Results table. For each domain:  number of objects $\# O$ in training instance, number of non-static predicates $\# P$, number of features to test $\# F$, and
		for each  input (full and partial graphs, plain traces): avg. number of admissible features $\# F_a$, avg. number of edges in input graphs $\#E$ (trace length for traces), avg. total time (data generation, learning, verification), and ratio of successful verification tests (Verif). Averages over 25 runs except for full graphs
		(not sampled)}
\label{table:results}
\end{table*}

\section{Experiments}

We have tested \sift   over a number of benchmarks. 
For this, a set of traces $T$ is  generated from one or more instances of a hidden domain $D$,
a domain $D_T$ is learned, and $D_T$ is verified over traces $T'$ from larger domain instances.
The experiments have been run on two types of Intel(R) Xeon(R) nodes: Platinum 8352M CPU,  running at 2.30GHz,
and  Gold 6330 CPU, running at 2.00GHz, using 22 cores. The code and  data are public \cite{goesgensjansen:siftcodeupload}.

\medskip

\noindent \textbf{Domains:} include Blocks with 3 and 4 operators, Delivery, Driverlog, Grid, Ferry, Gripper, Hanoi, Logistics,
Miconic, Sliding $n$-puzzle, Sokoban. These are all standard domains, some of which have been used in prior work on lifted model learning.
Grid-Lock and Sokoban-Pull are variations of  Grid and Sokoban, each one  adding one action schema to
make the resulting domains \emph{dead-end free} (an extra  lock action in Grid, and  a pull-action in Sokoban).
Dead-ends present a problem for data generation, as most random traces end up being trapped
in parts of the state space.

\medskip

\noindent \textbf{Training Data:} For each domain,  the training  data (traces) is obtained from a single large instance $P$
with approximately 100k edges.\footnote{Traces from multiple instances could have been used too.}
This size  is used to determine  the number of objects in the instance $P$ used to generate the data, except
for Logistics that required more data.  The (plain) traces are sampled
using two parameters: the number of traces $n$,  and their  length $L$ (number of actions). 
The first  trace is   sampled  from  the initial state $s_0$, while the rest are
sampled starting  in a state $s$  that is reached from $s_0$ in $m$ random steps, $2L \leq m \leq 5L$.
The number $n$ of traces has been set to $5$, and the length  of the traces have been set roughly to
the minimum  lengths $L$  needed so that 5 random traces of length $L$ result in learned
domains with $100\%$ validation success rates (as explained below). In some domains, the length $L$
required involves  tens of actions, in others, a few hundred. These are all plain traces with no state equalities.
A \emph{second type of training input} is considered for reference which uses full state graphs $G(P)$
as in  \cite{bonet:ecai2020}. This input  corresponds  to  traces augmented with state equalities.
The SAT approach can deal with graphs with a few hundred states; as we will see, \sift  can deal with hundreds of thousands,
without making assumptions that the graph is complete or that different nodes represent different states.
Indeed, a \emph{third type of training input} is considered as well which is given by a subgraph of $G(P)$. For this, a breadth first search
is done in $P$ from the initial state and a few other sampled states until  a number of states and edges are  generated that
yield   $100\%$ validation success rates. The difference with the plain traces, is that these are  extended traces
(state equalities) sampled in ``breadth'' and not in ``depth''. The number of sampled initial states is 5 except for Delivery, Driverlog, and Logistics
that required more data (resp. 10, 30, and 20 samples). In  Grid and Sokoban, a single large sample drawn from the initial state of $P$ was used
instead, because samples from deeper states fail to reach many other states,  as mentioned above.

\medskip

\noindent \textbf{Validation and Verification:} For testing validity and generalization, we consider hidden instances $P'=\tup{D,I'}$
that are larger than the instances $P=\tup{D,I}$ used in training, and use the methods above for obtaining  a set of traces 
and  of extended traces (partial state graphs) $T'$ from $P'$. For checking if these sets of validation and test  traces
is  compatible with the learned domain $D_T$, we check if there is  a node in the input where  an action $a(o)$ is done  in $T'$
such that a precondition $f(o')$  of $a(o)$ in $D_T$ is found to  be false in $n$.
If there is no such node, then $T'$ is regarded as being \emph{compatible} with  $D_T$. The reason that we cannot demand
the preconditions of $a(o)$ to be  true rather than being  ``not false'', is that the  information in a trace is incomplete.
We also test if sequences of actions that are not applicable in $P'$  are not applicable in the learned domain $D_T$ either.
For doing this in a sound manner,   if $\tau$ is trace from $P'$ that involves an action $a(o)$, we look for prefixes $\tau'$ of $\tau$
such that in the end node of $\tau'$, the action $a(o)$ is not applicable. Since, the action $a(o)$ occurs in $\tau$, the truth value
of all its preconditions $f(o')$ is known  in all such nodes, and hence if $a(o)$ is not applicable after the sequence $\tau'$ in $D$, it should not be applicable
in $D_T$ either (as in Theorem~\ref{thm:main}). We thus check for false positives and false negatives,
and report the \emph{percentage of successful verification tests (verification rate)}.  Both  plain and extended traces (partial graphs) $T'$ are used in these tests.

\medskip

\noindent \textbf{Results:} Table~\ref{table:results} shows the results. The first columns state the domain, the number of objects $\#O$ in the instance used
for training, the number of non-static predicates $\#P$ in the domain (hidden features), and the number of features $\#F$ whose consistency must be tested.
There are then three sets of columns for each of the three learning inputs: full state graphs, partial state graphs, and plain traces. Each includes
columns for number of features $\#F_a$ found to be consistent,  number of edges $\#E$ in the input graphs (for traces,  $\#E$ is  their length $L$),
and overall time, which includes data generation, feature generation and  pruning, and testing (verification rate;  Verif).
The results for  traces and partial state graphs  are \emph{averages} over 25 runs (\sift is a deterministic algorithm, but sampling is random).
As it can be seen from  the table, domains that verify  100\% of the test  traces are found in \emph{all the domains}
when the inputs are full state graphs and partial state graphs.
Plain traces  fail in  Grid and Sokoban (see below). 
The scalability of \sift  shows when processing full state
graphs with 500,000 edges or more, as  in the $n$-puzzle and Logistics.
In almost all domains, the number of admissible features learned $\#F_a$ is
slightly larger than the number of (non-static) predicates $\#P$, meaning that some ``redundant''
predicates from  $D_{max}$  are learned along with the hidden predicates in $D$.

Grid and Sokoban are not learned correctly from random traces, even if  Grid-Lock and Sokoban-Pull,  which contain all of their action
schemas, are. The reason is that the extra actions remove the dead-ends which prevent random traces
from being  sufficiently informative. The difficulty in Grid and Sokoban is thus not learning but sampling. 
Interestingly,  Grid also  illustrates  a case in which \emph{plain traces are strictly less informative than extended traces}. Indeed, one can show that
\emph{Grid cannot be learned correctly from plain traces alone}. The reason is that in any trace, a locked cell is opened once  from a specific cell,
and  there is  no way to determine  that the  same state can be reached by opening the lock from another cell.
Extended traces can represent state equality and avoid this problem, while plain traces do not cause this problem in Grid-Lock,
where a cell can be locked again and reopened from a different adjacent cell.

\medskip

\noindent \textbf{Analysis.} We have also looked at  the domains learned,  and they all look correct indeed,
containing the hidden predicates in  $D$, and  redundant predicates from $D_{max}$.
For example, in Blocksworld,  an undirected version of the \emph{on}  relation is learned  that is true if two  blocks are directly on
top of each other without revealing which one is above. In Gripper, an extra unary predicate is learned that captures
if a ball is being held, without specifying the gripper.
In  Driverlog, a learned predicate  keeps  track of the driver location even if on  a truck.
In Logistics, eight  predicates are learned, including a ternary relation that tracks both the location and city of a truck.
In the $n$-sliding puzzle, \sift  learns  24 ``redundant'' but meaningful features (details in the extended version).
All these predicates  are correct but redundant, and roughly  correspond to \emph{derived predicates} 
that can be tracked with action effects, and  hence, which do not need to be tracked by axioms.
They are all part of the maximal domain description $D_{max}$.

\section{Discussion}

We have presented the first general, and scalable solution to the problem
of learning lifted STRIPS models from action traces alone. The approach makes
use of the intuitions that underlie the LOCM systems \cite{locm1,locm}
but the formulation, the scope, and the theoretical guarantees
are  different. The learning task is challenging because
there is no information about the structure of states, which must
be fully inferred from the traces. The new approach is based on the notion
of \emph{features} $f=\tup{k,B}$ that represent  \emph{assumptions:}
the possibility of an  atom  $f(x)$ in the hidden domain of arity $k$, 
being affected by the action patterns in $B$ only.
The consistency of these assumptions can be tested efficiently over the
input traces $T$ by collecting a set $C_f(T)$ of tractable, 2-CNF-like
equality and inequality pattern constraints. The consistent features
define the learned domain which is guaranteed to generalize correctly
for a suitable finite set  of traces. 
The experiments show the generality and scalability of the approach.
Three direct extensions that we have not addressed in the paper
are: the the elimination of ``redundant'' features and predicates,
the derivation of static predicates of lower arity, 
and the variations needed  to make the learning approach robust to noisy inputs. For this,
notice that rather than  ``pruning'' a feature $f$ when found to be inconsistent with a
trace, $f$ can be pruned when inconsistent with $k$ traces.

\section*{Acknowledgments}

We thank Patrik Haslum for useful comments.
The research has been supported by the Alexander von Humboldt Foundation with funds from the German Federal Ministry for Education and Research.
This project has received funding from the European Research Council (ERC) under the European Union's Horizon 2020 research and innovations programme (Grant agreement No. 885107).
This project was also funded by the German Federal Ministry of Education and Research (BMBF) and the Ministry of Culture and Science of the German State of North Rhine-Westphalia (MKW) under the Excellence Strategy of the Federal Government and the L\"ander.

\bibliography{paper}

\section{Appendix}
\vspace{1\baselineskip}

\begin{proof}[Proof Theorem 9]
  In the main text.
\end{proof}

\begin{proof}[Proof Theorem 12]
	The truth values of all ground atoms are determined because the signs of the action preconditions and effects
	in $D$ are known,  and every ground atom $p(o)$ in $A(D,T)$ is,  by definition,  a precondition or effect of an action in $T$.
	Since the traces are connected the values propagate through them.
\end{proof}

\begin{proof}[Proof Theorem 15]
 The preconditions of the learned actions in the $D_T$ are all true in every node of the traces where the action is applied, so
that   the traces $T$ from $D$ are applicable in $D_T$. At the same time, $D_T$ contains predicates $f$ that represent  the (dynamic)
 predicates $p$  in $D$ (Theorem 17), as the corresponding features  are consistent with any set of traces from $D$,  and $T$ is complete.
 This means that $p$ and $f$ will have the same arity and involve the same action patterns. 
  Thus, if $p(o')$ is a precondition of the action $a(o)$ in $D$, then $f(o')$ will be a precondition of the same action in $D_T$ (possibly with the
  inverse sign),  and if $a(o)$ is not applicable after $\tau$ in $P$ because a $p(o')$-precondition of $a(o)$ is false, then
  $a(o)$ will  not be  applicable after $\tau$ in $P_T$ because the $f(o')$-precondition of $a(o)$ will also be false.
\end{proof}


\begin{proof}[Proof Theorem 17] For any set of traces $T$ drawn from $D$, the domain features $f$ will be consistent with $T$, as the actual signs of the action patterns
  $a[t]$ in $B_f$ that follow from $D$, provide a valuation of the action pattern signs $sign(a[t])$ that satisfy the pattern constraints $C_f(T)$.
  If $f(t[x])$ is an effect of action $a(x)$ in $D$, the sign of the action pattern $a[t]$ in $B_f$ is the sign of effect $f(t[x])$.
\end{proof}

\begin{proof}[Proof Theorem 19]
  Since $D_{max}$ is equivalent to $D$ in the sense of generating the same traces and graphs, then the proof argument for Theorem~17 applies here as well.
\end{proof}

\begin{proof}[Proof Theorem 20]
	A  feature $f$ that is not valid can be shown to be inconsistent in some set of extended traces.
	There is a finite set of features given a hidden domain, as the number of patterns only depends on the action arity.
	Thus only a finite set of finite extended traces is needed to rule out all invalid features.
\end{proof}

\begin{table*}[t]\setlength{\tabcolsep}{6.4pt}\centering
	\begin{tabular}{lrrrrrrrr}
		           &             \multicolumn{2}{c}{1 Trace}& \multicolumn{2}{c}{2 Traces}& \multicolumn{2}{c}{3 Traces}& \multicolumn{2}{c}{4 Traces}\\
		Domain     &  \#$F_a$ & Verif &  \#$F_a$ & Verif &  \#$F_a$ & Verif &  \#$F_a$ & Verif \\
		\midrule
		blocks3      & $6.6$&$ 68\%$& $5.1$&$ 96\%$& $5.0$&$100\%$& $5.0$&$100\%$\\
		blocks4      & $9.0$&$100\%$& $9.0$&$100\%$& $9.0$&$100\%$& $9.0$&$100\%$\\
		delivery     & $6.2$&$ 36\%$& $5.4$&$ 68\%$& $5.2$&$ 84\%$& $5.0$&$100\%$\\
		driverlog    &$13.9$&$ 44\%$&$13.0$&$100\%$&$13.0$&$ 96\%$&$13.0$&$100\%$\\
		ferry        & $4.0$&$100\%$& $4.0$&$100\%$& $4.0$&$100\%$& $4.0$&$100\%$\\
		grid\_lock   & $8.9$&$ 40\%$& $7.2$&$ 88\%$& $7.0$&$ 92\%$& $7.1$&$ 96\%$\\
		gripper      & $6.0$&$100\%$& $6.0$&$100\%$& $6.0$&$100\%$& $6.0$&$100\%$\\
		hanoi        & $4.9$&$ 84\%$& $4.6$&$ 92\%$& $4.0$&$100\%$& $4.0$&$100\%$\\
		logistics    & $7.8$&$ 44\%$& $7.2$&$ 88\%$& $7.2$&$ 80\%$& $7.0$&$ 96\%$\\
		miconic      & $8.0$&$100\%$& $8.0$&$100\%$& $8.0$&$100\%$& $8.0$&$100\%$\\
		npuzzle      &$28.2$&$ 64\%$&$27.2$&$ 84\%$&$26.3$&$ 92\%$&$26.0$&$100\%$\\
		sokoban\_pull& $4.6$&$ 92\%$& $3.0$&$100\%$& $3.0$&$100\%$& $3.0$&$100\%$\\
		
	\end{tabular}
	\caption{Results data when 1,2,3, or 4 traces considered instead of the 5 traces used in the paper. These can be regarded as ``intermediate'' results, although
     each is an  average over new traces.}
\end{table*}

\begin{table*}[t]
\setlength{\tabcolsep}{2.1pt}\centering\footnotesize
\begin{tabular}{llr@{}r@{}rl}
Domain        & Instance                                                    & (&\# V,&\# E) & Verification Instance               			\\
\midrule
blocks3       & $6$ blocks                                                  & $($&$  4051,$&$  21300)$           & $7$ blocks                                        			\\
blocks4       & $7$ blocks                                                  & $($&$ 65990,$&$ 186578)$           & $8$ blocks                                        			\\
delivery      & $3 \times 3$ grid $2$ packages $2$ trucks                   & $($&$  9639,$&$  57888)$           & $3 \times 3$ grid $3$ packages $2$ trucks             		\\
driverlog     & $5$ loc $2$ drivers $2$ trucks $2$ packages                 & $($&$ 10575,$&$  63720)$            & $7$ loc $2$ drivers $2$ trucks $3$ packages 	    		\\
ferry         & $5$ loc $5$ cars                                            & $($&$ 31250,$&$ 156250)$            & $6$ cars $5$ loc                            	     		\\
grid          & $3 \times 3$ grid $3$ keys $2$ shapes ($3$ locks)           & $($&$ 32967,$&$  99863)$    		 & $3 \times 4$ grid $4$ keys $2$ shapes ($6$ locks)     		\\
grid\_lock    & $3 \times 3$ grid $3$ keys $2$ shapes ($3$ locks)           & $($&$ 51436,$&$ 152040)$    		 & $3 \times 4$ grid $4$ keys $2$ shapes ($6$ locks)     		\\
gripper       & $2$ rooms $3$ grippers $7$ balls                            & $($&$ 17728,$&$  95680)$           & $2$ rooms $3$ grippers $8$ balls                             \\
hanoi         & $3$ pegs $9$ discs                                          & $($&$ 19683,$&$  59046)$           & $3$ pegs $10$ discs                               			\\
logistics     & $2$ plane $4$ truck $7$ loc $3$ city $2$ package            & $($&$ 54756,$&$ 648648)$           & $2$ plane $3$ truck $9$ loc $3$ city $2$ package         	\\
miconic       & $5$ floors $5$ persons                                      & $($&$ 38880,$&$ 127008)$           & $6$ floors $6$ persons                            			\\
npuzzle       & $3 \times 3$ grid $8$ tiles                                 & $($&$181440,$&$ 483840)$           & $4 \times 4$ grid $15$ tiles                      			\\
sokoban       & $4 \times 4$ grid ($4$ boxes)                               & $($&$ 10071,$&$  26834)$           & $5 \times 5$ grid ($3$ boxes)                     			\\
sokoban\_pull & $4 \times 4$ grid ($4$ boxes)                               & $($&$ 21824,$&$  66328)$           & $5 \times 5$ grid ($3$ boxes)                     			\\
\end{tabular}
\caption{Further details about the instances used in the experiments, including the number of nodes and edges in the full state graphs}
\end{table*}



\begin{table*}[tp]
	\centering\footnotesize\renewcommand{\arraystretch}{0.5}
	\begin{tabular}{l|ll|l}
		Feature&\multicolumn{2}{l|}{Patterns}&Meaning\\
		\midrule
		1
		&
		move-to-table$[1]$, &
		move-from-table$[1]$ & block is on table \\
		
		\midrule
		
		2 & move$[3]$, & move-from-table$[2]$ & block is clear \\
		& move$[2]$, & move-to-table$[2]$ &  \\
		
		\midrule
	
		3 & 
		move$[1,2]$, &  move-to-table$[1,2]$ & block at index 1 is stacked onto block at index 2 \\
		& move$[1,3]$, &  move-from-table$[1,2]$ &  \\
		
		\midrule
		4 &
		move-from-table$[2,1]$, & move$[3,1]$ & block at index 2 is stacked onto block at index 1 \\
		&	move$[2,1]$, & move-to-table$[2,1]$ &  \\
		
		\midrule
		5 &
		move$[1,2]$, & move-to-table$[1,2]$ & blocks at indexes 1 and 2 are stacked onto each other\\
		& move$[2,1]$, & move-to-table$[2,1]$ & \\
		& move-from-table$[2,1]$, & move$[3,1]$ &  \\
		& move$[1,3]$, & move-from-table$[1,2]$  &

	\end{tabular}
	\caption{List of admissible features for blocks4. Actions are move among blocks, and move to and from table. Left patterns are positive (adds), and right patterns are negative (deletes)}
\end{table*}

\begin{table*}[tp]
	\centering\footnotesize\renewcommand{\arraystretch}{0.5}
	\begin{tabular}{l|ll|l}
		Feature&\multicolumn{2}{l|}{Patterns}&Meaning\\
		\midrule
		1 & 
		pick$[1]$, & drop$[1]$ & ball is grabbed \\
		
		\midrule
		
		2 & 
		move$[1]$, & move$[2]$ & location of the robot \\
		
		\midrule
		
		3 & 
		drop$[3]$, & pick$[3]$ & is gripper free \\
		
		\midrule
		
		4 & 
		pick$[1,2]$, & drop$[1,2]$ & location of ball \\
		
		\midrule
		
		5 & 
		move$[2,1]$, & move$[1,2]$ & location with previous room \\
		
		\midrule
		
		6 & 
		pick$[3,1]$, & drop$[3,1]$ & grabbed ball in gripper \\

	\end{tabular}
	\caption{List of admissible features for gripper}
\end{table*}

\begin{table*}[tp]
	\centering\footnotesize\renewcommand{\arraystretch}{0.5}
	\begin{tabular}{l|ll|l}
		Feature&\multicolumn{2}{l|}{Patterns}&Meaning\\
		\midrule

		1 & 
		unload$[1]$, & load$[1]$ & package is loaded \\
		
		\midrule
		
		2 & 
		drive$[1, 2]$, & drive$[1, 3]$ & location of truck \\
		
		\midrule
		
		3 & 
		fly$[1, 2]$, & fly$[1, 3]$ & location of plane \\
		
		\midrule
		
		4 & 
		unload$[2, 1]$, & load$[2, 1]$ & location of package \\
		
		\midrule
		
		5 & 
		drive$[1, 2]$, & drive$[1, 3]$ & location of vehicle \\
		&fly$[1, 2]$, & fly$[1, 3]$ &\\
		
		\midrule
		
		6 & 
		load$[1, 3]$, & unload$[1, 3]$ & package in vehicle \\
		
		\midrule
		
		7 & 
		drive$[3, 4, 1]$, & drive$[2, 4, 1]$ & location of truck in city \\

	\end{tabular}
	\caption{List of admissible features for logistics}
\end{table*}

\begin{table*}[tp]
	\centering\footnotesize\renewcommand{\arraystretch}{0.5}
	\begin{tabular}{l|ll|l}
		Feature&\multicolumn{2}{l|}{Patterns}&Meaning\\
		\midrule

		1 & 
		disembark-truck$[1]$, & board-truck$[1]$ & Driver $<1>$ is currently inside a truck \\
		
		\midrule
		
		2 & 
		load-truck$[1]$, & unload-truck$[1]$ & package $<1>$ is loaded \\
		
		\midrule
		
		3 & 
		disembark-truck$[2]$, & board-truck$[2]$ & Driver seat of Truck $<2>$ is empty \\
		
		\midrule
		
		4 & 
		drive-truck$[4, 3]$, & board-truck$[1, 3]$ & location of the driver while driving \\
		&disembark-truck$[1, 3]$, & drive-truck$[4, 2]$&\\
		
		\midrule
		
		5 & 
		walk$[1, 2]$ & drive-truck$[4, 2]$& location of the driver\\
		&drive-truck$[4, 3]$, & walk$[1, 3]$& independent of driving status \\
		
		\midrule
		
		6 &
		walk$[1, 2]$, & board-truck$[1, 3]$ & location of the driver while walking \\
		&disembark-truck$[1, 3]$, & walk$[1, 3]$&\\
		
		\midrule
		
		7 & 
		unload-truck$[1, 3]$, & load-truck$[1, 3]$ & location of package \\
		
		\midrule
		
		8 & 
		drive-truck$[1, 2]$, & drive-truck$[1, 3]$ & location of truck \\
		
		\midrule
		
		9 & 
		disembark-truck$[2, 3]$, & board-truck$[2, 3]$ & truck parked at location \\
		
		\midrule
		
		10 & 
		drive-truck$[1, 2]$, & disembark-truck$[2, 3]$ & location of truck while driving \\
		&drive-truck$[1, 3]$, & board-truck$[2, 3]$ & \\
		
		\midrule
		
		11 & 
		disembark-truck', (1, 2), & board-truck', (1, 2) & driver driving truck \\
		
		\midrule
		
		12 & 
		load-truck', (1, 2), & unload-truck', (1, 2) & package in truck \\
		
		\midrule
		
		13 & 
		drive-truck$[4, 1, 2]$, & disembark-truck$[1, 2, 3]$ & location of truck driver pair while driving \\
		&board-truck$[1, 2, 3]$,& drive-truck$[4, 1, 3]$&\\
		
	\end{tabular}
	\caption{List of admissible features for driverlog}
\end{table*}


\begin{table*}[p]\footnotesize\renewcommand{\arraystretch}{0.5}
	\centering
\begin{tabular}{l|ll|l}
	Feature&\multicolumn{2}{l|}{Patterns}&Meaning\\
	\midrule
	1
	&
	move-down$[2, 4]$, & move-up$[2, 4]$, 
	&blank has has a different x,y coordinate.\\
	&move-left$[4, 3]$, & move-right$[4, 3]$,
	&Negated Hidden Domain predicate.\\
	&move-down$[2, 3]$, & move-up$[2, 3]$,
	&\\
	&move-left$[2, 3]$, & move-right$[2, 3]$
	&\\
	\midrule
	
	2
	&
	move-up$[1, 3, 2]$, & move-down$[1, 3, 2]$,
	&Tile $<1>$ is not on this y,x coordinate.\\
	&move-right$[1, 3, 2]$, & move-left$[1, 3, 2]$,
	&Negated Hidden Domain predicate.\\
	&move-up$[1, 4, 2]$, & move-down$[1, 4, 2]$,
	&\\
	&move-right$[1, 3, 4]$, & move-left$[1, 3, 4]$
	&\\
	\midrule
	
	3
	&
	move-left$[4]$,&
	move-right$[2]$
	&blank is right of me.\\
	\midrule
	
	4
	&
	move-down$[4]$,&
	move-up$[3]$
	&blank is above of me.\\
	\midrule
	
	5
	&
	move-right$[1, 2]$,&
	move-left$[1, 4]$
	&Tile $<1>$is right of me.\\
	\midrule
	
	6
	&
	move-up$[4, 1]$,&
	move-down$[3, 1]$
	&Tile $<1>$is not below of me.\\
	\midrule
	
	7
	&
	move-right$[4]$,&
	move-left$[2]$
	&blank is left of me.\\
	\midrule
	
	8
	&
	move-up$[4]$,&
	move-down$[3]$
	&blank is below of me.\\
	\midrule
	
	9
	&
	move-left$[1, 2]$,&
	move-right$[1, 4]$
	&Tile $<1>$is left of me.\\
	\midrule
	
	10
	&
	move-down$[4, 1]$,&
	move-up$[3, 1]$
	&Tile $<1>$is not above of me.\\
	\midrule
	
	11
	&
	move-left$[4]$, & move-right$[4]$,
	&black has a different x corrdinate.\\
	&move-right$[2]$, & move-left$[2]$
	&\\
	\midrule
	
	12
	&
	move-right$[1, 2]$, & move-left$[1, 2]$,
	&Tile $<1>$has a different x corrdinate.\\
	&move-right$[1, 4]$, & move-left$[1, 4]$
	&\\
	\midrule
	
	13
	&
	move-up$[4]$, & move-down$[4]$,
	&blank has a different y coordinate.\\
	&move-down$[3]$, & move-up$[3]$
	&\\
	\midrule
	
	14
	&
	move-down$[3, 1]$, & move-up$[3, 1]$,
	&Tile $<1>$has a different y corrdinate.\\
	&move-down$[4, 1]$, & move-up$[4, 1]$
	&\\
	\midrule
	
	15
	&
	move-right$[4, 2]$,&
	move-left$[2, 4]$
	&Arrow pointing to blank (left)\\
	&&&if blank is left.\\
	\midrule
	
	16
	&
	move-down$[4, 3]$,&
	move-up$[3, 4]$
	&Arrow pointing to blank (up)\\
	&&&if blank is above.\\
	\midrule
	
	17
	&
	move-right$[2, 1, 4]$,&
	move-left$[4, 1, 2]$
	&Arrow pointing to Tile $<1>$(right)\\
	&&&if Tile $<1>$is right.\\
	\midrule
	
	18
	&
	move-down$[3, 1, 4]$,&
	move-up$[4, 1, 3]$
	&Arrow pointing to Tile $<1>$(down)\\
	&&&if Tile $<1>$is below.\\
	\midrule
	
	19
	&
	move-right$[2, 4]$,&
	move-left$[4, 2]$
	&Arrow pointing from blank (right)\\
	&&&if blank is left.\\
	\midrule
	
	20
	&
	move-down$[3, 4]$,&
	move-up$[4, 3]$
	&Arrow pointing from blank (down)\\
	&&&if blank is above.\\
	\midrule
	
	21
	&
	move-right$[4, 1, 2]$,&
	move-left$[2, 1, 4]$
	&Arrow pointing from Tile $<1>$(left)\\
	&&&if Tile $<1>$is right.\\
	\midrule
	
	22
	&
	move-down$[4, 1, 3]$,&
	move-up$[3, 1, 4]$
	&Arrow pointing from Tile $<1>$(up)\\
	&&&if Tile $<1>$is below.\\
	\midrule
	
	23
	&
	move-right$[4, 2]$, & move-right$[2, 4]$,
	&Undirected x-Edge if blank is left.\\
	&move-left$[4, 2]$, & move-left$[2, 4]$
	&Or Arrow pointing from blank on x axis.\\
	\midrule
	
	24
	&
	move-down$[4, 3]$, & move-down$[3, 4]$,
	&Undirected y-Edge if blank is above.\\
	&move-up$[4, 3]$, & move-up$[3, 4]$
	&Or Arrow pointing from blank on y axis.\\
	\midrule
	
	25
	&
	move-right$[2, 1, 4]$, & move-right$[4, 1, 2]$,
	&Undirected edge on x-axis if Tile $<1>$is right.\\
	&move-left$[4, 1, 2]$, & move-left$[2, 1, 4]$
	&Or Arrow pointing to Tile $<1>$on x axis.\\
	\midrule
	
	26
	&
	move-up$[4, 1, 3]$, & move-up$[3, 1, 4]$,
	&Undirected edge on y-axis if Tile $<1>$is above.\\
	&move-down$[4, 1, 3]$, & move-down$[3, 1, 4]$
	&Arrow pointing from Tile $<1>$on y axis.\\
	
\end{tabular}
\caption{List of admissible features for npuzzle}
\end{table*}

\end{document}